\documentclass[10pt,twocolumn,letterpaper]{article}

\usepackage{iccv}
\usepackage{times}
\usepackage{graphicx}
\usepackage{amsmath}
\usepackage{amssymb}
\usepackage[dvipsnames]{xcolor}
\usepackage{subcaption}
\usepackage{adjustbox}
\usepackage{algorithm}
\usepackage{algpseudocode}
\usepackage{mathtools}

\usepackage[pagebackref,breaklinks,colorlinks]{hyperref}

\DeclareMathOperator{\sign}{sign}
\DeclareMathOperator{\clamp}{clamp}
\def\-{\raisebox{.75pt}{-}}

\newcommand{\comment}[1]{}

\iccvfinalcopy 

\ificcvfinal\fi

\begin{document}

\title{Deep 3D-to-2D Watermarking: Embedding Messages in 3D Meshes and Extracting Them from 2D Renderings}

\author{Innfarn Yoo\textsuperscript{1}\hspace{1.0em}Huiwen Chang\textsuperscript{1}\hspace{1.0em}Xiyang Luo\textsuperscript{1}\hspace{1.0em}Ondrej Stava\textsuperscript{2}\\
Ce Liu\textsuperscript{1}\thanks{Currently affiliated with Microsoft Azure AI.}\hspace{0.35em}\hspace{1.0em}Peyman Milanfar\textsuperscript{1}\hspace{1.0em}Feng Yang\textsuperscript{1}\\
\textsuperscript{1}Google Research\hspace{1em}\textsuperscript{2}Google LLC\\
{\tt\small \{innfarn, huiwenchang, xyluo, ostava, milanfar, fengyang\}@google.com\hspace{1em}ce.liu@microsoft.com}}
\maketitle
\ificcvfinal\thispagestyle{empty}\fi

\begin{abstract}
Digital watermarking is widely used for copyright protection. Traditional 3D watermarking approaches or commercial software are typically designed to embed messages into 3D meshes, and later retrieve the messages directly from distorted/undistorted watermarked 3D meshes. However, in many cases, users only have access to rendered 2D images instead of 3D meshes. Unfortunately, retrieving messages from 2D renderings of 3D meshes is still challenging and underexplored. We introduce a novel end-to-end learning framework to solve this problem through: 1) an encoder to covertly embed messages in both mesh geometry and textures; 2) a differentiable renderer to render watermarked 3D objects from different camera angles and under varied lighting conditions; 3) a decoder to recover the messages from 2D rendered images. From our experiments, we show that our model can learn to embed information visually imperceptible to humans, and to retrieve the embedded information from 2D renderings that undergo 3D distortions. In addition, we demonstrate that our method can also work with other renderers, such as ray tracers and real-time renderers with and without fine-tuning.
\end{abstract}

\section{Introduction}
\begin{figure}[t]
  \centering
  \includegraphics[width=\linewidth]{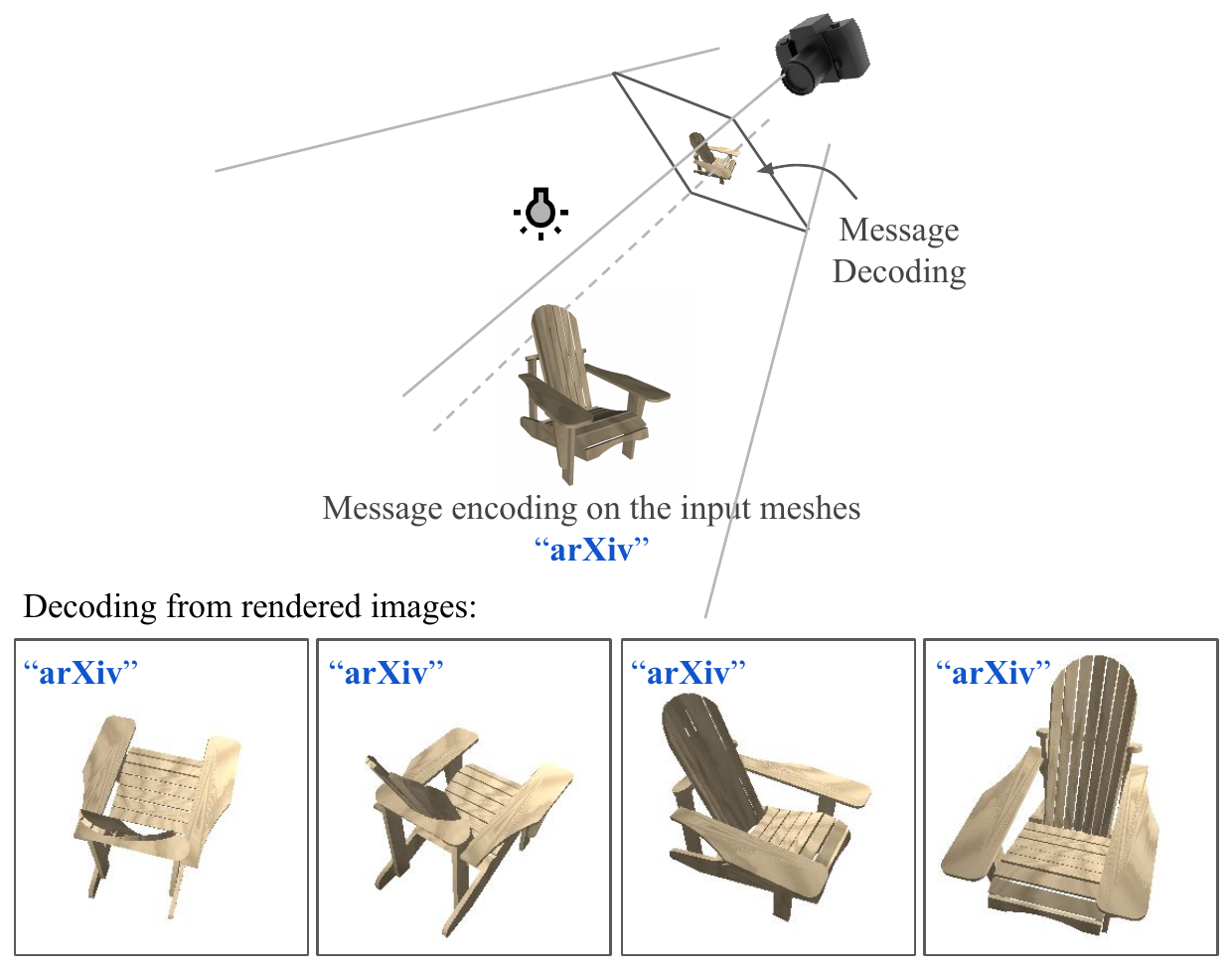}
  \caption{A chair mesh is watermarked with a message ``arXiv", then rendered as images. The embedded message can be retrieved from rendered images from different views and lighting conditions.}
  \label{fig:teaser}
\end{figure}

\noindent Digital watermarking is a key technology for copyright protection, source tracking, and authentication for digital content.
The goal of digital watermarking is to embed messages in another media, \emph{e.g.}, image, video, or 3D, and be able to decode the messages even after the watermarked media is copied, edited or distorted. There are two types of watermarking: invisible or visible. In this paper, we will mainly focus on invisible watermarking which means the watermarked media should be perceptually the same as the original media.

\begin{figure*}[ht]
  \centering
  \includegraphics[width=1.0\linewidth]{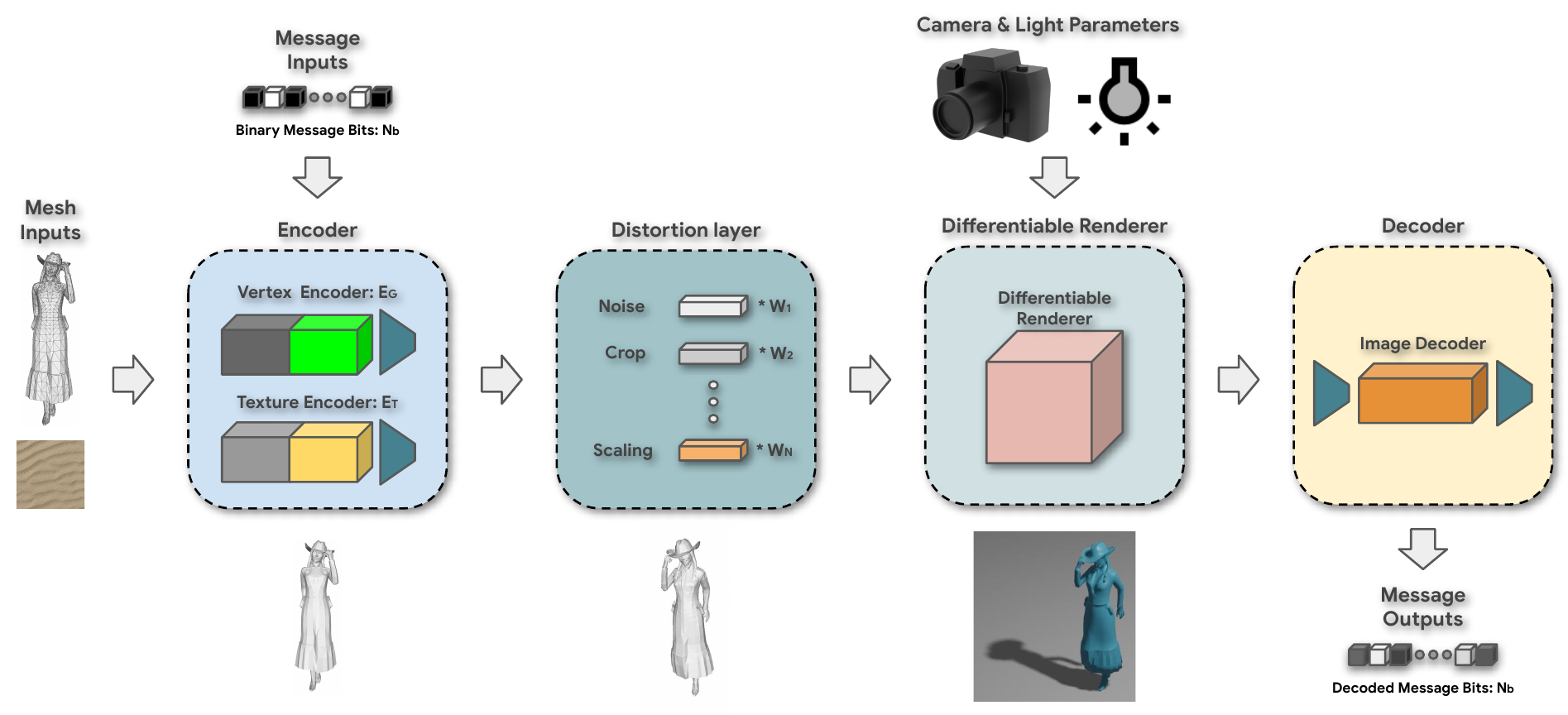}
  \caption{Overview of our deep watermarking pipeline. The encoder embeds messages into 3D meshes in an imperceptible fashion. Then we apply some distortions like crop, scaling to the watermarked 3D mesh. The rendered images are generated using a differentiable renderer. After that we extract the messages from the rendered 2D images.}
  \label{fig:pipeline}
\end{figure*}

3D models, due to their increasing ubiquity in a wide range of applications such as movie making, gaming, 3D printing, augmented reality (AR), 3D mapping, \emph{etc.}, have become an important subject of digital watermarking. \emph{The classic problem of watermarking for 3D models has been primarily formulated as embedding and retrieving  messages, both in 3D space}. While traditional 3D watermarking methods~\cite{10.1145/311535.311540,4694850} and commercial tools~\cite{watermark3d} are useful in 3D manufacturing and printing industries, they are inapplicable to mediums such as gaming, movie making and graphic design, where \emph{the messages need to be retrievable from 2D renderings of the 3D models}.

In this work, we study the problem of 3D-to-2D watermarking -- how to embed messages invisibly in 3D meshes and recover them robustly from 2D renderings. Recently, a number of research works~\cite{ReDMark,Luo_2020_CVPR,HiDDeN} shows that approaches based on deep neural networks can achieve the state-of-the-art performance on image watermarking. One possible solution that takes advantage of these methods is to embed and retrieve messages in the rendered 2D images of 3D objects.
However, this approach cannot be easily adapted to decode 2D renderings of the same objects from different camera views or under varied lighting and shading conditions and thus is impractical. Another solution is to embed messages in the 2D textures of 3D meshes and extract the messages from 2D renderings. Our experiments show that it does not work since the existing deep image watermarking solution is not robust to 3D to 2D rendering distortion. 
To address these issues, we propose an end-to-end trainable deep 3D-to-2D watermarking framework which is resistant to various 3D distortions and capable of extracting messages from images rendered under different lighting conditions and camera views. 

Our framework consists of 1) an \textit{encoder}; 2) a \textit{distortion layer}; 3) a \textit{differentiable renderer}; and 4) a \textit{decoder} as shown in Fig.~\ref{fig:pipeline}. 
Specifically, inspired by the advances of 3D reconstruction using differentiable renderers ~\cite{liu2019soft,drcTulsiani17,yifan2019differentiable}, we employ a state-of-the-art differentiable renderer to bridge the gap between the 3D encoding and 2D decoding stages.
In the encoding stage, our model first embeds message bits into either 3D geometry and/or 2D textures of the original 3D data. In the decoding stage, it learns to extract message bits from 2D images generated by the differentiable renderer. 
We build a 3D-to-2D watermarking benchmark including $5146$ 3D objects with textures, and we investigate the performance of different architectures on it. We then analyze the performance of our method by measuring its capacity,~\emph{i.e.}, the size of the message we can embed, its robustness,~\emph{i.e.}, the bit accuracy with respect to various distortions, and its quality or invisibility,~\emph{i.e.}, difference between watermarked/unwatermarked 2D renderings. Finally, we show that our model can work with other renderers, such as ray tracing and real-time renderers, as well as the availability of fine-tuning our decoder for other not differentiable renderers.

Our key contributions are following:
\begin{itemize}
\item To the best of our knowledge, this paper presents the first 3D-to-2D watermarking method which can retrieve messages encoded in 3D meshes from its rendered 2D image and broaden 3D watermarking usage.
\item The use of differentiable rendering makes our method fully-differentiable, which enables to train the whole framework end-to-end with a collection of differentiable 3D distortions.
\item Our decoder can decode the embedded messages from non-differentiable renderers, and can be improved further by fine-tuning.
\end{itemize}

While our model performs well as shown in Sec.~\ref{sec:experiments}, it is still limited in multiple aspects. For example, our method has low bit capacities compared to traditional 3D watermarking methods. However, low bit capacities are still useful for copyright protection (~\emph{e.g.}, zero-bit watermarking). Another limitation is that if attackers use totally different style rendering techniques such as cartoon rendering, our model needs to be re-trained.
\section{Related Work}
\label{sec:related_work}

\noindent Methods relevant to our work are categorized as watermarking, differentiable rendering, neural network architectures, and others.

\subsection{Watermarking}
\label{ssec:prev_watermarking}

\noindent There is a vast body of work related to watermarking on 3D visual data. Several papers~\cite{4412893,4694850} gave a nice survey on 3D watermarking.
Early 3D watermarking approaches~\cite{doi:10.1111/1467-8659.t01-1-00597,10.1145/311535.311540,10.1145/1022431.1022456} leveraged Fourier or wavelet analysis on triangular or polygonal meshes.
Recently, Hou~\emph{et al.}~\cite{7954665} introduced a 3D watermarking method using the layering artifacts in 3D printed objects.
Son~\emph{et al.}~\cite{10.1007/978-981-10-4154-9_37} used mesh saliency as a perceptual metric to minimize vertex distortions.
Hamidi~\emph{et al.}~\cite{info10020067} further extended mesh saliency with wavelet transform to make 3D watermarking robust.
Jing~\emph{et al.}~\cite{doi:10.1177/1550147719826042} studied watermarking for point clouds through analyzing vertex curvatures.

Deep learning based methods for image watermarking~\cite{ReDMark,Luo_2020_CVPR,ROMark,SteganoGAN,HiDDeN} achieved great progress in recent years. HiDDeN~\cite{HiDDeN} was one of the first deep image watermarking methods that achieved good performance compared to traditional watermarking approaches. Many extensions of this work have since been proposed, including a novel architecture based on circulant convolutions~\cite{ReDMark}, improvement of the robustness for more complex distortions~\cite{Luo_2020_CVPR,tancik2019stegastamp,wengrowski2019light}, and a novel two-stage framework for complex distortions~\cite{10.1145/3343031.3351025}.

\subsection{Differentiable Rendering}
\label{ssec:prev_differentiable_rendering}

\noindent Differentiable rendering has been primarily developed for explicit 3D representations such as meshes, point cloud, and volume \cite{liu2019soft,qi2017pointnet,sitzmann2019deepvoxels,yifan2019differentiable}. While explicit 3D representations are commonly used due to their ease of manipulation and popularity in the downstream applications and 3D software, they have historically been precluded from deep learning pipelines due to the difficulty in back-propagating through mesh rasterization. 
Loper~\emph{et al.}~\cite{loper2014opendr} first proposed a general differentiable renderer named OpenDR, which can efficiently approximate derivatives with respect to 3D models. Later, SoftRasterizer~\cite{liu2019soft}, DiffRen~\cite{genova2018unsupervised} and Differentiable Ray-tracer~\cite{li2018differentiable} have emerged and devised different techniques to make rendering back-propagatable. In this paper, we use 3D mesh representation to be consistent with conventional 3D watermarking work~\cite{doi:10.1111/1467-8659.t01-1-00597,10.1145/311535.311540,10.1145/1022431.1022456,4694850}, and we employ DiffRen~\cite{genova2018unsupervised} in our differentiable rendering component.

\subsection{Neural Network Architectures}
\label{ssec:prev_nn_arch}

\noindent For 3D data, PointNet~\cite{PointNet} and PointNet++~\cite{PointNet2} used multiple shared multi-layer perceptrons (MLPs) with transform layers, and have been widely adopted to extract features from 3D point clouds. The PointNet architectures can be easily extended to 3D mesh vertices.
Convolutional mesh autoencoders~\cite{Ranjan_2018_ECCV} applied Chebyshev convolution filters on their autoencoder architecture as well as barycentric mesh resampling.
A 3D variational autoencoder, which handles both mesh connectivity and geometry, was introduced in~\cite{Tan_2018_CVPR}. Recently, a fully convolutional mesh autoencoder~\cite{zhou2020fully} allows variable-size mesh inputs.

In the 2D image domain, standard convolutional neural networks (CNN)~\cite{AlexNet} have shown significant progress on classifying objects, and have been aggressively extended to many other areas. The accuracy of image classification and object detection is further improved by deeper networks.
U-Net~\cite{UNet} symmetrically connects encoder and decoder feature maps in variational autoencoder architectures, and shows improvements on image segmentation tasks.

\subsection{Other Related Work}
\label{ssec:prev_other_related_work}

\noindent Evaluating perceptual quality of 3D meshes is an open problem with no standardized metrics.
For many applications, using simple metrics such as Hausdorff distance~\cite{Hausdorff} or root mean square error is sufficient, but these metrics tend to correlate poorly with the actual quality perceived by human vision.
Visual quality measures~\cite{SGP:SGP03:042-051} used an additional smoothness factor that correlates better with human vision than pure spatial measures.
The same concept was further improved in~\cite{doi:10.1111/j.1467-8659.2011.02017.x}, where a multi-scale mesh quality metric was proposed to rely on curvature statistics on different neighborhood scales.
\emph{The limitations of these methods are non-differentiability and their reliance on purely 3D geometric data without taking into consideration other factors such as materials, textures or lighting.}
In \cite{7272102}, it was shown that using image-based quality assessment on rendered images of the 3D models is a viable strategy that can implicitly handle all surface and material properties for evaluation.
\section{Method}
\label{sec:method}

\noindent Fig.~\ref{fig:pipeline} shows our deep 3D watermarking pipeline, which consists of four modules: a) an encoder; b) a distortion layer; c) a differentiable renderer; and d) a decoder. We will go deep into each module in the following sections.

\subsection{Definitions}
\noindent We denote by $\mathbb{M}(V, F, T, P)$ the input mesh. The vertices $V \in \mathbb{R}^{N_v \times C_v}$ contain $N_v$ vertices and each vertex has $C_v$  vertex elements such as 3D position, normal, vertex color, and 2D texture coordinate elements. The mesh faces $F \in \{0, \cdots, N_v-1\}^{N_f \times C_f}$ contains vertex indices, where $N_f$ is the number of faces and $C_f$ is the number of face indices with value $3$ for a triangular mesh and $4$ for a quadrilateral mesh. $T\in\mathbb{R}^{H_t \times W_t \times C_t}$ store texture information, where $W_t$, $H_t$ and $C_t$ are width, height, number of color channels of the texture respectively. The mesh material color information $P \in \mathbb{R}^{10}$ includes ambient, diffuse, specular RGB color, and a shininess. We denote by $M \in \{0,1\}^{N_b}$ the binary message with length $N_b$ to be embedded into the mesh.

\subsection{Encoder}
\label{ssec:encoders}

\begin{figure*}[ht]
  \centering
  \begin{subfigure}{0.24\linewidth}
    \centering
    \includegraphics[width=1.0\linewidth]{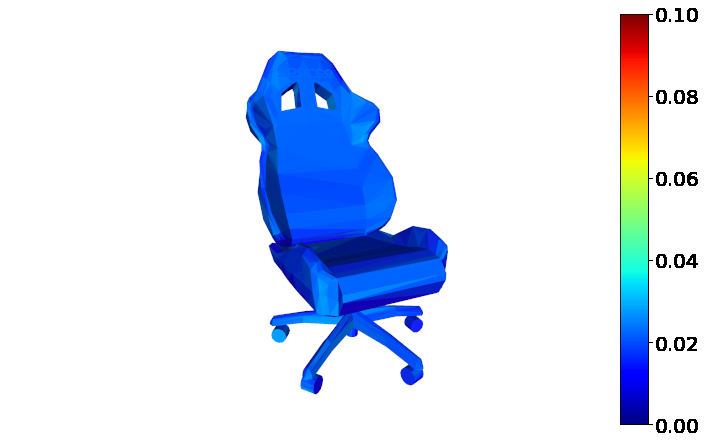}
  \end{subfigure}
  \begin{subfigure}{0.24\linewidth}
    \centering
    \includegraphics[width=1.0\linewidth]{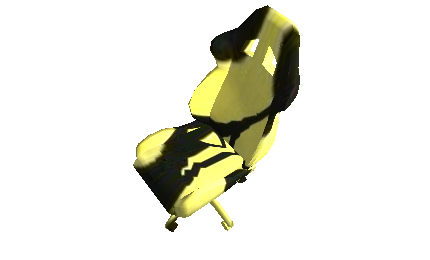}
  \end{subfigure}
  \begin{subfigure}{0.24\linewidth}
    \centering
    \includegraphics[width=1.0\linewidth]{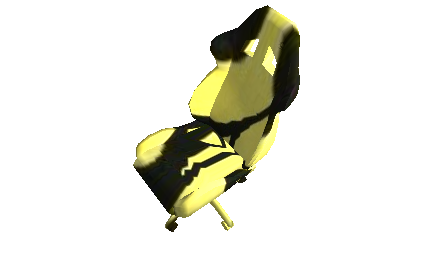}
  \end{subfigure}
  \begin{subfigure}{0.24\linewidth}
    \centering
    \includegraphics[width=1.0\linewidth]{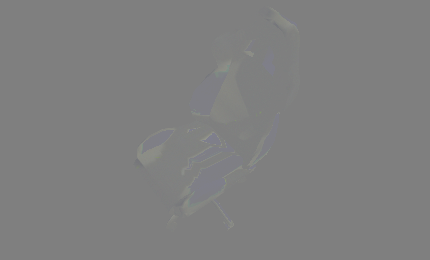}
  \end{subfigure}
  \\
  \begin{subfigure}{0.24\linewidth}
    \centering
    \includegraphics[width=1.0\linewidth]{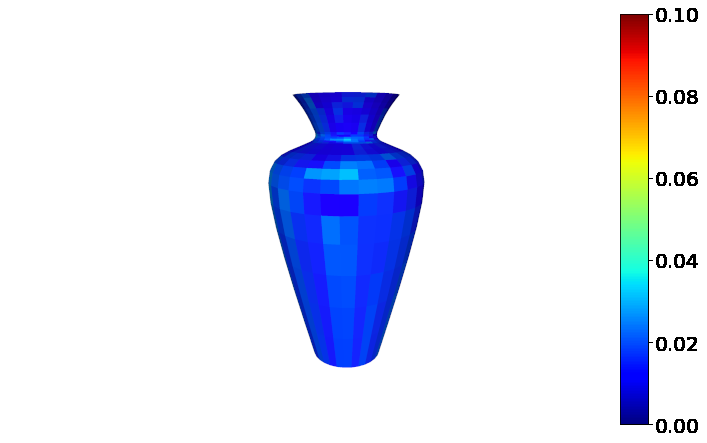}
    \caption{Vertex Difference}
  \end{subfigure}
  \begin{subfigure}{0.24\linewidth}
    \centering
    \includegraphics[width=1.0\linewidth]{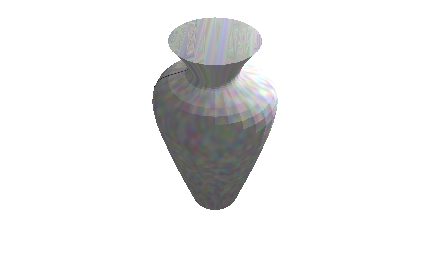}
    \caption{Not watermarked}
  \end{subfigure}
  \begin{subfigure}{0.24\linewidth}
    \centering
    \includegraphics[width=1.0\linewidth]{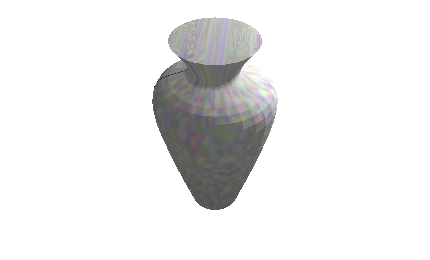}
    \caption{Watermarked}
  \end{subfigure}
  \begin{subfigure}{0.24\linewidth}
    \centering
    \includegraphics[width=1.0\linewidth]{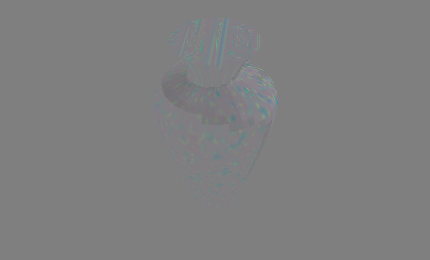}
    \caption{Difference (4x)}
  \end{subfigure}
  \caption{The differences of geometric elements of a mesh are color-coded and shown in (a). The rendered image from input mesh, the rendered image from watermarked mesh, and the (4x) difference between input and watermarked images in (b), (c), and (d) respectively.}
  \label{fig:watermarked_mesh}
\end{figure*}

\begin{figure}[h]
  \centering
  \begin{subfigure}{0.32\linewidth}
    \centering
    \includegraphics[width=1.0\linewidth]{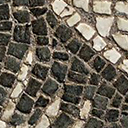}
  \end{subfigure}
  \begin{subfigure}{0.32\linewidth}
    \centering
    \includegraphics[width=1.0\linewidth]{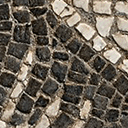}
  \end{subfigure}
  \begin{subfigure}{0.32\linewidth}
    \centering
    \includegraphics[width=1.0\linewidth]{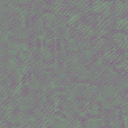}
  \end{subfigure}
  \\
  \begin{subfigure}{0.32\linewidth}
    \centering
    \includegraphics[width=1.0\linewidth]{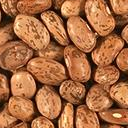}
    \caption{Not watermarked}
  \end{subfigure}
  \begin{subfigure}{0.32\linewidth}
    \centering
    \includegraphics[width=1.0\linewidth]{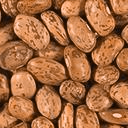}
    \caption{Watermarked}
  \end{subfigure}
  \begin{subfigure}{0.32\linewidth}
    \centering
    \includegraphics[width=1.0\linewidth]{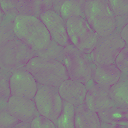}
    \caption{Difference (4x)}
  \end{subfigure}
  \caption{The input texture, the watermarked texture, and the (4x) difference images are shown in (a), (b), and (c) respectively.}
  \label{fig:watermarked_texture}
\end{figure}

\noindent Changing mesh faces $F$ produces undesirable artifacts and the material color information $P$ is too small to hide information. Thus, we decide to embed message in vertex and texture components. Note that we only encode message into normal and texture coordinate elements in the vertex, since changing vertex position creates degenerated triangles and prohibits backpropagation. We could embed message in either component or both of them. At the encoder, we replicate each message bit $N_v$ times to construct a tensor with dimension $N_v \times N_b$ to embed message in the vertex component. To embed the message in texture, we replicate each message bit $H_t \times W_t$ times to construct a tensor with dimension $H_t \times W_t \times N_b$. After this, we concatenate the message tensors with input vertices and/or textures. We define this concatenated tensor as $V_m \in \mathbb{R}^{N_v \times (C_v + N_b)}$ and $T_m \in \mathbb{R}^{H_t \times W_t \times (C_t + N_b)}$ for vertices case and texture case respectively. Then we get watermarked vertices $V_e = E_G(V_m) \in \mathbb{R}^{N_v \times C_v} $ and watermarked texture $T_e = E_T(T_m) \in\mathbb{R}^{H_t \times W_t \times C_t}$, where $E_G$ and $E_T$ are vertex and texture message embedding neural network respectively.
For $E_G$, we made the PointNet architecture~\cite{PointNet} to be fully-convolutional, and also used a CNN-based architecture for $E_T$ similar to~\cite{HiDDeN}.
Fig.~\ref{fig:watermarked_mesh} and Fig.~\ref{fig:watermarked_texture} show the comparisons of watermarked meshes and textures. Visually we could not tell the difference between watermarked/non-watermarked pairs.

\subsection{Distortions}
\label{ssec:distortions}
\noindent To make our watermarking system robust to different distortions, we add a distortion layer in our watermarking training pipeline. Several commonly used distortions are considered: 1) additive Gaussian noise with mean $\mu$ and standard deviation $\sigma$; 2) random axis-angle rotation with parameters ($\alpha$, $x$, $y$, $z$); 3) random scaling with a parameter $s$; and 4) random cropping on \emph{3D vertices}. Since all these distortions are differentiable, we could train our network end-to-end.

\subsection{Differentiable Rendering}
\label{ssec:DN_rendering}
\begin{figure}[h]
  \centering
  \begin{subfigure}{.15\linewidth}
    \centering
    Real Texture
  \end{subfigure}
  \begin{subfigure}{0.25\linewidth}
    \centering
    \frame{\includegraphics[width=1.0\linewidth]{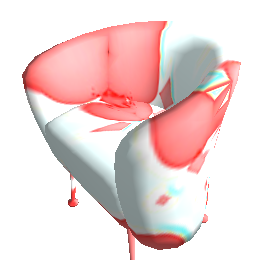}}
  \end{subfigure}
  \begin{subfigure}{0.25\linewidth}
    \centering
    \frame{\includegraphics[width=1.0\linewidth]{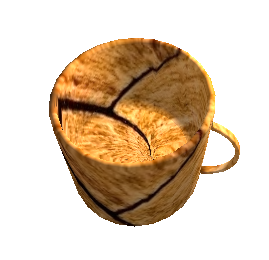}}
  \end{subfigure}
  \begin{subfigure}{0.25\linewidth}
    \centering
    \frame{\includegraphics[width=1.0\linewidth]{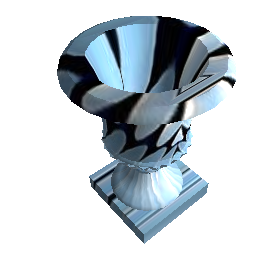}}
  \end{subfigure}
  \\
  \begin{subfigure}{.15\linewidth}
    \centering
    Noise Texture
  \end{subfigure}
  \begin{subfigure}{0.25\linewidth}
    \centering
    \centering
    \frame{\includegraphics[width=1.0\linewidth]{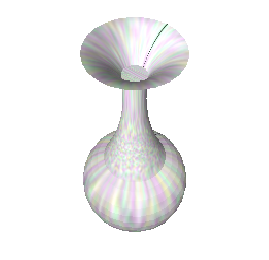}}
  \end{subfigure}
  \begin{subfigure}{0.25\linewidth}
    \centering
    \centering
    
    \frame{\includegraphics[width=1.0\linewidth]{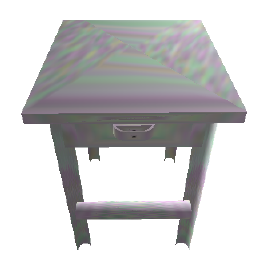}}
  \end{subfigure}
   \begin{subfigure}{0.25\linewidth}
    \centering
    \centering
    
    \frame{\includegraphics[width=1.0\linewidth]{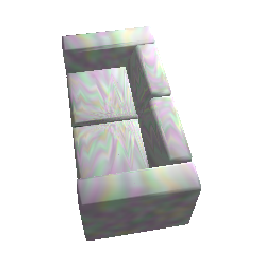}}
  \end{subfigure}
  \caption{Examples of differentiable rendering results.}
  \label{fig:diffren_results}
\end{figure}

\noindent To train a 3D watermarking system that can extract messages from the 2D rendered images, we need a differentiable rendering layer.
We leverage the state-of-the-art work in the field of differentiable rendering and follow the work from Genova~\emph{et al.}~\cite{genova2018unsupervised}, which proceeds in three steps:
1) rasterization, which computes screen-space buffers per pixel and the barycentric coordinates of the pixel inside triangles; 2) deferred shading; 3) splatting, where each rasterized surface point is converted into a splat, centered at the pixel and colored by the corresponding shaded color. The approach computes smooth derivatives on the pixel grid w.r.t the mesh vertices and per-vertex attributes. In this paper, we use the Phong reflection model~\cite{phong1975illumination} for shading.
Our differentiable rendering examples are shown in Fig.~\ref{fig:diffren_results}.

In our pipeline, we assume there are a fixed number of point light sources positioned at $L^i_P$ with intensity $L^i_I$. Given the lighting parameters $L$, camera matrices $K$, and meshes $\mathbb{M}$, we use our differentiable renderer to generate the output 2D images,
\begin{equation}
    I = R_D(\mathbb{M}, K, L) \in \mathbb{R}^{H_r \times W_r \times 3}.
\end{equation}

\subsection{Decoder}
\label{ssec:decoders}
\noindent We use a neural network $D$ to retrieve a message $M_r$ from the rendered image $I$, ~\emph{i.e.}, $M_r=D(I)$. The decoder network $D$ uses a global pooling layer after several convolution layers to allow variable size input images, and then several fully-connected layers. The last fully-connected layer has a fixed number of output nodes which is the length of the message (more details are in the supplementary material).
To further improve the decoder performance on non-differentiable renderers such as commercial software~\cite{Blender}, we could fine-tune or train the decoder alone with the specific rendered output. We will show a specific example,~\emph{i.e.}, real-time and physically-based renderers in Sec.~\ref{ssec:decoding_from_other_rendering}.

The final message bits $M_{rb} \in \{0, 1\}^{N_b}$ are from the binarization of the predicted message with Eq.~\ref{eq:binarization}.
\begin{equation}
    \label{eq:binarization}
    M_{rb} = \clamp(\sign(M_{r} - 0.5), 0, 1).
\end{equation}
Note that $M_r$ is used for computing the message loss in Sec.~\ref{ssec:losses}, and $M_{rb}$ is used for evaluating bit accuracy.


\subsection{Losses}
\label{ssec:losses}

\noindent We model the objective of 3D-to-2D watermarking by optimizing: 1) the vertex watermarking loss, 2) the texture watermarking loss, 3) the 2D rendering loss, and 4) the message loss. 

Among them, the vertex watermarking loss computes the distance between the watermarked mesh vertices $V_e$ and the original mesh vertices $V$,
\begin{equation}
    L_{vertex}(V, V_e) = \sum_{i}{w^i\frac{\sum_{\alpha}|V^i[\alpha] - V^i_e[\alpha]|}{N_v C_v} },
    \label{eq:geo_loss}
\end{equation}
where $i$ indicates a component of vertices, such as normal or texture coordinates, and $w^i$ is the weight for each component. By adjusting $w_i$, we can adjust the sensitivity of change for each vertex component. Similarly, texture loss computes the difference between the watermarked texture $T_e$ and the original texture $T$,
\begin{equation}
    \begin{split}
        L_{texture}(T, T_e)  = \frac{\sum_{\alpha} |T[\alpha] - T_e[\alpha]|}{H_t  W_t  C_t}.
    \end{split}
    \label{eq:texture_loss}
\end{equation}

\noindent The 2D rendering loss calculates the difference between the rendered images $I_o$ of the original meshes and the ones $I_w$ of watermarked mesh 
\begin{equation}
    \begin{split}
        L_{image}(I_o, I_w)  = \frac{\sum_{\alpha} |I_o[\alpha] - I_w[\alpha]|}{ 3H_w W_w}.
    \end{split}
    \label{eq:rendered_image_loss}
\end{equation}

\noindent The message loss penalizes the decoding error, which is defined by the difference between the predicted message $M_{rb}$ and the ground truth message $M$,
\begin{equation}
    \begin{split}
        L_{message}(M, M_r)  = \frac{\sum_{\alpha} |M[\alpha] - M_r[\alpha]|}{N_b}.
    \end{split}
    \label{eq:message_loss}
\end{equation}

\noindent Finally, the total loss is the weighted sum of all losses,
\begin{equation}
\begin{split}
    L_{total} = & \lambda L_{vertex} + \gamma L_{texture} + \\
    & \delta L_{image} + \theta L_{message} + \eta L_{reg},
\end{split}
\label{eq:total_loss}
\end{equation}
where $L_{reg}$ is the regularization loss, ~\emph{i.e.}, the $L_2$ sum of all network weights, $\lambda$, $\gamma$, $\delta$, $\theta$, and $\eta$ are hyper-parameters.

\section{Experiments}
\label{sec:experiments}

\subsection{Dataset}
\label{ssec:dataset_and_preprocessing}

\begin{figure}[h]
  \centering
  \includegraphics[width=1.0\linewidth]{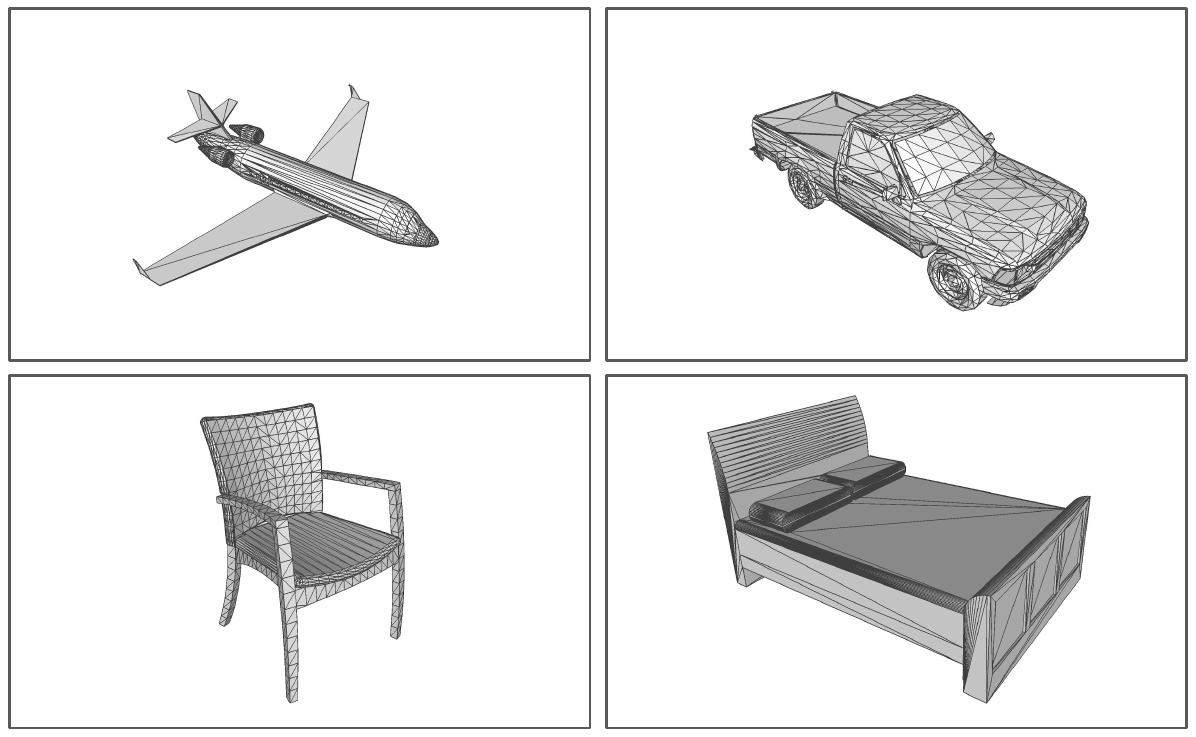}
  \caption{Examples of input (simplified) meshes.}
  \label{fig:ex_input_meshes}
\end{figure}

\noindent We use ModelNet 40-class dataset~\cite{Wu_2015_CVPR} for training and testing.
We normalize the input vertices' position elements $\{x, y, z\}$ to $[-1, 1]$, while keeping ratios between width, height and depth.
Then, the input meshes are simplified to fix the number of triangles and vertices using the CGAL library~\cite{cgal} as shown in Fig.~\ref{fig:ex_input_meshes}.
Fixing the number of vertices and triangles in the dataset is for training efficiency and memory limitation.
But our network could deal with meshes with different sizes by applying a global pooling layer in the vertex encoder architecture (the details are discussed in the supplementary material).
We also manually filter out low quality simplified meshes.
As a result, the number of meshes in ModelNet 40-class dataset is reduced to $5146$ meshes for training and $1329$ meshes for testing. Note that the original dataset has $9843$ and $2468$ meshes for training and testing, respectively.
ModelNet dataset does not contain texture information, thus we allocate texture coordinates using spherical mapping and get texture from either Gaussian smoothing a white noise or randomly selecting and cropping a texture from an open texture dataset~\cite{zhou2018nonstationary}.

\begin{figure*}[h]
  \centering
  \includegraphics[width=1.0\linewidth]{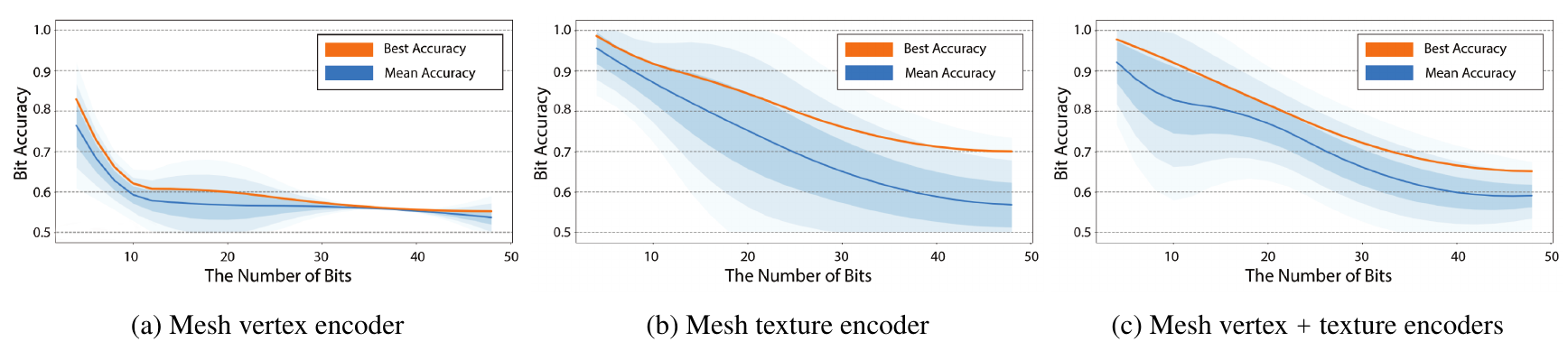}
  \begin{center}
  \caption{The relationship between bit accuracy and the length of message. We trained our pipeline 10 times with the different number of bits, then calculated best and mean accuracies as well as the standard deviations.}
  \label{fig:graph_acc_bits}
  \end{center}
\end{figure*}

\subsection{Implementation Details}
\label{ssec:hyper_params}
\noindent \textbf{Mesh parameters} The input meshes have $N_v = 5000$, $N_f = 5000$, $C_f = 3$, and $C_v = 5$. The input texture is with height $128$, width $128$ and number of channels $3$.
The rendered image is with width $600$ and height $400$.

\noindent \textbf{Training and testing parameters} For evaluation of our method, we randomly choose $100$ meshes in the test dataset, and take the average of output values. We use batch size $4$, learning rate $0.0001$ for training. And also we set the weights in the total loss function as $\lambda = 2.0$, $\gamma = 1.0$, $\delta = 1.0$, $\theta \in [0.1, 2.0]$, and $\eta = 0.01$. For deciding hyper-parameters, we sweep broad ranges of hyper-parameters and manually choose the best.

\noindent \textbf{Rendering parameters} We use common camera parameters, which include camera location, camera look-at location, camera up, and field of view ($fov_y$).
We use a point light source, the parameters are a point light location, colors, and attenuation factors.
While training, camera and lighting parameters are \emph{randomly generated in the given ranges.}
In our experiment, we use \emph{random camera locations} with ranges, $x = 0$, $y \in [\-3, \-2]$, $z \in [2, 4]$. Camera is always looking at the object center with $fov_y = 60^{\circ}$.
\emph{The point light location is also randomly sampled} from a Gaussian distribution with mean $\mu = (2, 1, 2)$ and standard deviation $\sigma = 0.2$.

We put all the network architecture and training details in supplemental material. Note that the following experiments embed messages with 8 bits unless mentioned specifically.

\begin{table}[h]
\centering
\resizebox{\columnwidth}{!}{%
\begin{tabular}{c|c c c|c c|c c}
    \hline
             & \multicolumn{3}{c|}{Bit Accuracy} & \multicolumn{2}{c|}{Geometry $L_1$ Diff} & \multicolumn{2}{c}{Rendered Image} \\
    \hline
    Architectures & Best   & $\mu$  & $\sigma$ & Normal & Texcoord & PSNR & SSIM \\ \hline
    PointNet      & \textbf{0.8553} & 0.7591 & 0.0789 & 0.0529 & 0.0805 & \textbf{25.57} & \textbf{0.9425} \\
    PointNet v2   & 0.8207 & 0.7213 & 0.0754 & \textbf{0.0415} & \textbf{0.0745} & 25.10 & 0.9416 \\
    \hline
\end{tabular}}
\caption{Comparisons of vertex encoder architectures with a texture encoder.}
\label{tab:comp_3d_enc_archs_with_tex}
\end{table}

\subsection{Architectures}
\label{ssec:comp_arch}
\noindent We investigated different architectures for the vertex encoder, the texture encoder, and the image decoder.

For the vertex encoder, we compared two architectures: 1) PointNet; and 2) fully convolutional PointNet (noted PointNet v2). Tab.~\ref{tab:comp_3d_enc_archs_with_tex} and Tab.~\ref{tab:comp_3d_enc_archs_without_tex} show the comparisons with and without a texture encoder respectively.
As shown in the tables, both architectures showed similar performance on bit accuracy, geometry $L_1$ differences, and rendered image quantitative metrics, however, \emph{we chose the fully convolutional PointNet, since it can accept meshes of different sizes as an input at inference.}

\begin{table}[h]
\centering
\resizebox{\columnwidth}{!}{%
\begin{tabular}{c |c c c|c c|c c}
    \hline
             & \multicolumn{3}{c|}{Bit Accuracy} & \multicolumn{2}{c|}{Geometry $L_1$ Diff} & \multicolumn{2}{c}{Rendered Image} \\
    \hline
    Architectures & Best & $\mu$  & $\sigma$ & Normal & Texcoord & PSNR & SSIM \\ \hline
    PointNet  & \textbf{0.6837} & 0.6151 & 0.0366 & 0.1873 & 0.0546 & 28.47 & \textbf{0.9563} \\
    PointNet v2  & 0.6616 & 0.6255 & 0.0216 & \textbf{0.1506} & \textbf{0.0430} & \textbf{28.83} & 0.9557 \\
    \hline
\end{tabular}}
\caption{Comparisons of vertex encoder architectures without a texture encoder.}
\label{tab:comp_3d_enc_archs_without_tex}
\end{table}

For the texture encoder, we compare the network architecture used a CNN-based architecture~\cite{HiDDeN} and a fully convolutional U-Net architecture as shown in Tab.~\ref{tab:comp_texture_encoders}. U-Net got better bit accuracy while HiddeN generated less visible watermarked texture. Note that \cite{Luo_2020_CVPR} has the same encoder and decoder architecture with HiDDeN~\cite{HiDDeN}.

\begin{table}[h]
\centering
\resizebox{0.95\columnwidth}{!}{%
\begin{tabular}{c|c c c|c c c} 
    \hline
            & \multicolumn{3}{c|}{Bit Accuracy} & \multicolumn{3}{c}{Texture} \\
    \hline
    Architectures   & Best   & $\mu$  & $\sigma$ & $L_1$ Diff & PSNR  & SSIM \\ \hline
    CNN (HiDDeN)    & 0.8269 & 0.7614 & 0.0411 & \textbf{0.03536} & \textbf{26.20} & 0.9113 \\ 
    U-Net           & \textbf{0.8474} & 0.7451 & 0.1289 & 0.05692 & 23.55 & \textbf{0.9143} \\ \hline
\end{tabular}}
\caption{Comparisons of texture encoder architectures.}
\label{tab:comp_texture_encoders}
\end{table}

Lastly, for the decoder, we evaluated HiDDeN~\cite{HiDDeN} and two simple baselines -- a standard CNN (4 layers) and a residual CNN (4 layers).
As shown in Tab.~\ref{tab:comp_image_decoders}, all the architectures can detect the message and among them, the HiDDeN decoder, probably benefited from its larger model size and capacity, performs better than the other models.

\begin{table}[h]
\centering
\resizebox{0.72\columnwidth}{!}{%
\begin{tabular}{c|c c c}
    \hline
            & \multicolumn{3}{c}{Bit Accuracy} \\
    \hline
    Architectures & Best & $\mu$ & $\sigma$ \\ \hline
    Simple CNN (4-layers) & 0.6396 & 0.6184 & 0.0415 \\
    Residual CNN (4-layers) & 0.7390 & 0.6453 & 0.0332 \\
    CNN (HiDDeN) & \textbf{0.8269} & 0.7614 & 0.0411 \\
    \hline
\end{tabular}}
\caption{Comparisons of decoder architectures.}
\label{tab:comp_image_decoders}
\end{table}

\subsection{Bit Accuracy vs. Message Length}
\label{ssec:acc_bits}

\noindent We use message length as $N_b \in \{4, 8, 16, 32, 48\}$, and launch 10 experiments for each message length and show the relationship between bit accuracy and the length of message in Fig.~\ref{fig:graph_acc_bits}. The blue line and areas show the average bit accuracy, $1\sigma$, $2\sigma$, and $3\sigma$ ranges, respectively. We could see that the bit accuracy drops when the number of bits increases. 

\subsection{Message Embedding Strategies}
\label{ssec:mess_embedding_stragegies}
\noindent As shown in Tab.~\ref{tab:encoders_and_accs}, when using ``vertex only" encoder, the watermarked mesh can hold less bits compared to ``texture only" or ``vertex $+$ texture" encoder. Combining vertex and texture encoders shows better bit accuracy compared to the ``texture only" encoder in higher numbers of bits.
\begin{table*}[h]
\centering
\resizebox{1.9\columnwidth}{!}{%
\begin{tabular}{c|c c c|c c|c c c|c c c}
    \hline
             & \multicolumn{3}{c|}{Bit Accuracy}& \multicolumn{2}{c|}{Geometry $L_1$ Diff} & \multicolumn{3}{c|}{Texture Diff} & \multicolumn{3}{c}{Rendered} \\
    \hline
    $\theta$ & Best & $\mu$ & $\sigma$ & Normal & Texture & $L_1$ & PSNR & SSIM & $L_1$ & PSNR & SSIM \\
    \hline
    0.01 & 0.5475 & 0.4997 & 0.0090 & 0.0475 & \textbf{0.0537} & \textbf{0.0042} & \textbf{46.57} & \textbf{0.9849} & \textbf{0.0055} & \textbf{36.91} & \textbf{0.9700} \\
    0.1 & 0.7947 & 0.7256 & 0.0396 & 0.0472 & 0.0621 & 0.0056 & 41.11 & 0.9844 & 0.0062 & 35.67 & 0.9614 \\
    1.0 & 0.9262 & 0.8857 & 0.0255 & \textbf{0.0437} & 0.0699 & 0.0186 & 30.74 & 0.8639 & 0.0063 & 35.35 & 0.9609 \\
    2.0 & 0.9325 & \textbf{0.8921} & 0.0432 & 0.2071 & 0.2404 & 0.0229 & 29.37 & 0.8062 & 0.0071 & 34.70 & 0.9603 \\
    5.0 & 0.8775 & 0.8562 & 0.0146 & 0.2563 & 0.2873 & 0.0363 & 25.45 & 0.6626 & 0.0088 & 32.90 & 0.9552 \\
    10.0 & \textbf{0.9362} & 0.8288 & 0.1135 & 0.1041 & 0.1810 & 0.0549 & 22.42 & 0.5088 & 0.0105 & 31.10 & 0.9426 \\
    \hline
\end{tabular}}
\caption{The message loss weight shows a trade-off relationship with the differences between the input and watermarked meshes. Higher message loss weight provides higher bit accuracy, however the output watermarked meshes are changed more.}
\label{tab:message_loss_weights}
\end{table*}

\begin{table}[h]
\centering
\resizebox{1.0\columnwidth}{!}{%
\begin{tabular}{c|c c c c c}
    \hline
                     & 4 Bits & 8 Bits & 16 Bits & 32 Bits & 48 Bits \\ \hline
    Vertex Only      & 0.7646 & 0.6289 & 0.5717 & 0.5627 & 0.5370 \\
    Texture Only     & \textbf{0.9554} & \textbf{0.8970} & 0.7975 & 0.6364 & 0.5679 \\
    Vertex + Texture & 0.9208 & 0.8480 & \textbf{0.8000} & \textbf{0.6441} & \textbf{0.5905} \\
    \hline
\end{tabular}}
\caption{The comparisons of bit accuracies and different message embedding strategies. The accuracies are the averages of 10 experiments.}
\label{tab:encoders_and_accs}
\end{table}

\subsection{Bit Accuracy vs. Mesh Quality}
\label{ssec:bit_acc_vs_L1}

\noindent As shown in Tab.~\ref{tab:message_loss_weights}, there is a clear trade-off between bit accuracy and mesh quality.
Higher bit accuracy results in higher $L_1$ difference as well as lower PSNR and SSIM, both in textures and rendered images.
We could control this trade-off by changing the message loss weight $\theta$.

\subsection{Distortions}
\label{ssec:comp_distortions}
\begin{table}[h]
\centering
\resizebox{0.6\columnwidth}{!}{%
\begin{tabular}{c|c}
    \hline
    Distortion Type & Bit Accuracy \\ \hline
    No Distortion  & 0.9046 \\
    Noise ($\sigma=0.01$)  & 0.9036 \\
    Rotation ($\pm \pi / 6$) & 0.9028 \\
    Scaling ($< 25 \%$) & 0.8953 \\
    Cropping ($< 20 \%$) & 0.8840 \\
    \hline
\end{tabular}}
\caption{The effect of different distortions.}
\label{tab:comp_distortions}
\end{table}

\noindent We evaluated the robustness of our methods to different distortions for example noise, rotation, scaling and cropping in Tab.~\ref{tab:comp_distortions}. We could see that our method is quite robust to different distortions.
More details about the experiments above are discussed in the supplementary material.

\subsection{Comparison with Deep Image Watermarking}
\label{ssec:comp_deep_image_watermarking}
\noindent We also compared with a deep image watermarking method~\cite{HiDDeN}. To make a fair comparison, we trained HiDDeN encoder and decoder on our texture images. For our method, we used the trained HiDDeN image encoder as our texture encoder, fixed it during training, and only trained our vertex encoder and image decoder. After training our network, we generated the rendered images, and compared the performance of decoding message bits from the rendered images using HiDDeN decoder and our decoder.
As shown in Tab.~\ref{tab:comp_image_watermarking}, our decoder could successfully decode message bits while the other showed nearly random bits. This shows the effectiveness of our end-to-end trainable 3D-2D watermarking framework.

\begin{table}[h]
\centering
\resizebox{0.6\columnwidth}{!}{%
\begin{tabular}{c|c c}
    \hline
             & HiDDeN~\cite{HiDDeN} & Our Method \\ \hline
    Bit Acc  & 0.5132 & 0.8213 \\
    \hline
\end{tabular}}
\caption{The comparison with a deep image watermarking method.}
\label{tab:comp_image_watermarking}
\end{table}

\subsection{Decoding from Other Renderers}
\label{ssec:decoding_from_other_rendering}

\begin{figure}[ht]
  \centering
  \begin{subfigure}{0.32\linewidth}
    \centering
    \includegraphics[width=1.0\linewidth, trim=20mm 0 20mm 0, clip]{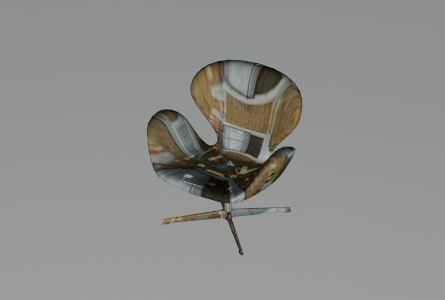}
  \end{subfigure}
  \begin{subfigure}{0.32\linewidth}
    \centering
    \includegraphics[width=1.0\linewidth, trim=20mm 0 20mm 0, clip]{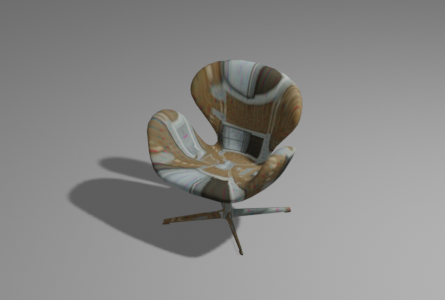}
  \end{subfigure}
  \begin{subfigure}{0.32\linewidth}
    \centering
    \includegraphics[width=1.0\linewidth, trim=20mm 0 20mm 0, clip]{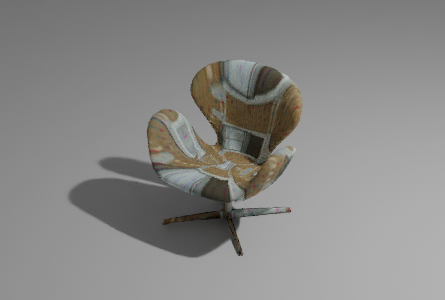}
  \end{subfigure}
  \\
  \begin{subfigure}{0.32\linewidth}
    \centering
    \includegraphics[width=1.0\linewidth, trim=20mm 0 20mm 0, clip]{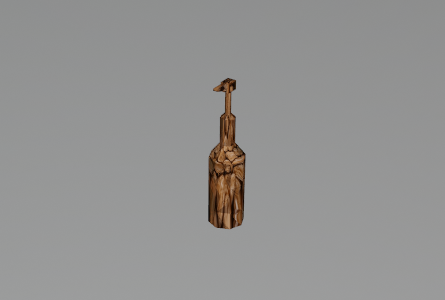}
    \caption{Workbench (RT)}
  \end{subfigure}
  \begin{subfigure}{0.32\linewidth}
    \centering
    \includegraphics[width=1.0\linewidth, trim=20mm 0 20mm 0, clip]{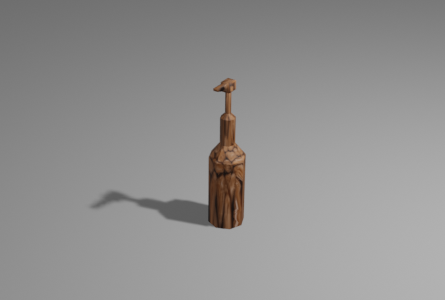}
    \caption{Eevee (PB)}
  \end{subfigure}
  \begin{subfigure}{0.32\linewidth}
    \centering
    \includegraphics[width=1.0\linewidth, trim=20mm 0 20mm 0, clip]{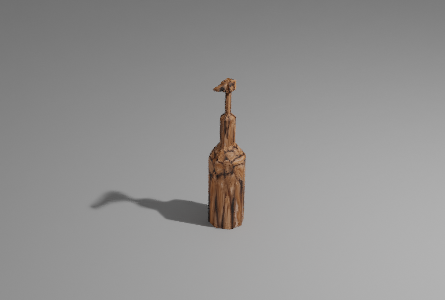}
    \caption{Cycle (PB)}
  \end{subfigure}
  \caption{Examples of different rendering outputs: (a) a RT renderer, (b) a PB renderer, and (c) a PB ray tracer.}
  \label{fig:different_renderer_output}
\end{figure}

\noindent In Tab.~\ref{tab:different_renderers}, row a-d show the bit accuracy of our decoder (without fine-tuning), decoding from the rendering results of multiple renderers. It includes our differentiable renderer, a real-time (RT) renderer, and two physically-based (PB) renderers. Note that although the bit accuracy is still low, it is not random. Our decoder could still recover some information. With technologies like channel coding, we could achieve higher bit accuracy by sacrificing some capacity, \emph{i.e.}, the number of useful bits we embedded.
Examples of rendering results are shown in Fig.~\ref{fig:different_renderer_output}.

\begin{figure}[ht]
  \centering
  \begin{subfigure}{0.32\linewidth}
    \centering
    \includegraphics[width=1.0\linewidth,trim=10mm 0 10mm 0, clip]{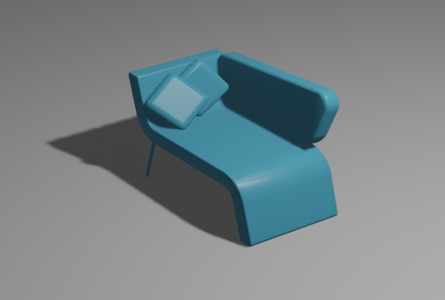}
  \end{subfigure}
  \begin{subfigure}{0.32\linewidth}
    \centering
    \includegraphics[width=1.0\linewidth,trim=10mm 0 10mm 0, clip]{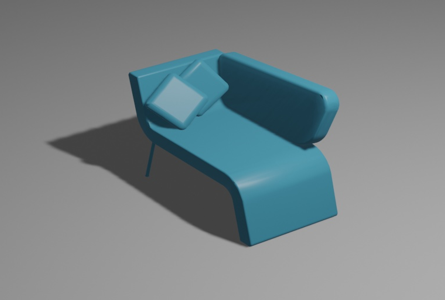}
  \end{subfigure}
  \begin{subfigure}{0.32\linewidth}
    \centering
    \includegraphics[width=1.0\linewidth,trim=10mm 0 10mm 0, clip]{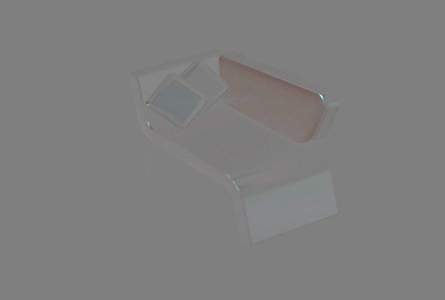}
  \end{subfigure}
  \\
  \begin{subfigure}{0.32\linewidth}
    \centering
    \includegraphics[width=1.0\linewidth, trim=20mm 5mm 20mm 5mm, clip]{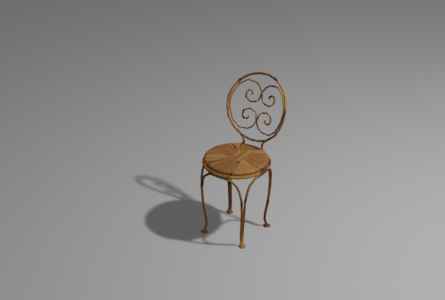}
    \caption{Not watermarked}
  \end{subfigure}
  \begin{subfigure}{0.32\linewidth}
    \centering
    \includegraphics[width=1.0\linewidth, trim=20mm 5mm 20mm 5mm, clip]{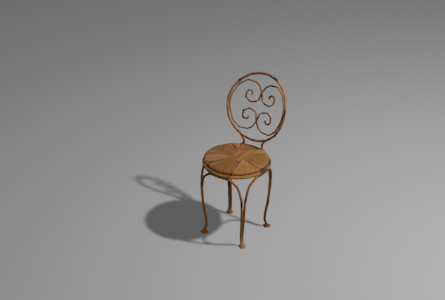}
    \caption{Watermarked}
  \end{subfigure}
  \begin{subfigure}{0.32\linewidth}
    \centering
    \includegraphics[width=1.0\linewidth, trim=20mm 5mm 20mm 5mm, clip]{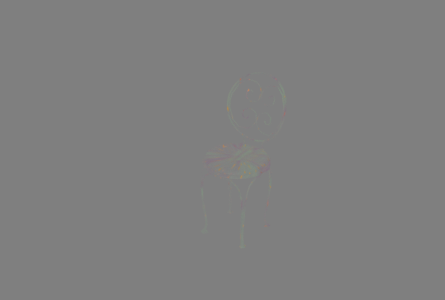}
    \caption{Difference (4x)}
  \end{subfigure}
  \caption{Examples of a ray traced image from an original mesh (a), a ray traced image from a watermarked mesh (b), and the (4x) difference image.}
  \label{fig:watermarking_ray_tracing}
\end{figure}

Moreover, though we cannot train our framework end-to-end for non-differentiable rendering methods, we can fine-tune our decoder with their rendering outputs to improve the bit accuracy as shown in Tab.~\ref{tab:different_renderers}e. Here we created a dataset of $6000$  watermarked meshes for training and $1500$ for testing, and then generated their rendering results from the Eevee renderer. Using the rendered images, we fine-tuned our decoder.
Fig.~\ref{fig:watermarking_ray_tracing} shows the ray traced images from an original mesh and a watermarked mesh as well as the difference between the two images.

\newcommand{\imageborder}{20mm}

\begin{table}[h]
\centering
\resizebox{0.65\columnwidth}{!}{%
\begin{tabular}{c|c|c}
    \hline
      & Renderer Type & Bit Acc \\ \hline
    a & Differentiable Renderer  & 0.7490 \\
    b & Workbench (RT) Renderer & 0.6341 \\
    c & Cycle (PB) Ray Tracer & 0.6634 \\
    d & Eevee (PB) Renderer  & 0.5973 \\ \hline
    e & Fine-tuned (Eevee) & 0.7679 \\
    \hline
\end{tabular}}
\caption{The accuracy of decoded message bits from different renderers including a RT renderer, PB renderers (a-d) by our default decoder, and a fine-tuned decoder (e).}
\label{tab:different_renderers}
\end{table}

\subsection{User Study on Rendered Images}
\label{ssec:user_study}
\noindent The quantitative metrics, such as $L_1$, PSNR, and SSIM, may not be able to fully reflect human perceptions. To assess whether users can distinguish original and watermarked images, we conduct two crowd-sourced user studies through Amazon's Mechanical Turk.
We show the original and watermarked renderings side-by-side, in shuffled left-and-right order.
We then ask participants to determine whether the two images are ``identical" or ``not identical" in a few seconds. A total of $200$ pairs in each case were rated, where every pair was independently rated by $5$ raters. 

Tab.~\ref{tab:user_study} shows that in 74\% of the noise-textured cases, participants responded ``identical". For the user study, we used the rendered images from our differentiable renderer.
For the ray traced images, in 78\% of the cases, participants responded ``identical". 

\begin{table}[h]
\centering
\resizebox{0.8\columnwidth}{!}{%
\begin{tabular}{c|c c c c}
    \hline
           & Same Rating & PSNR & SSIM & $L_1$ Diff \\
    \hline
    Texture & 74.07 \% & 47.76 & 0.997 & 0.0009 \\
    Ray Tracing & 78.06 \% & 54.52 & 0.999 & 0.0004 \\
    \hline
 \end{tabular}}
\caption{User study about watermarked and non watermarked images. Raters were asked whether the two images are  ``identical” or ``not identical”.}
\label{tab:user_study}
\end{table}

\section{Conclusion, Limitations, and Future Work}
\label{sec:conclusion}

\noindent We study the problem of 3D-to-2D watermarking, a key step towards an automatic and secure solution to protect copyright and integrity for 3D models from their 2D rendered images. Our contributions include the first end-to-end trainable model resistant to various 3D distortions and manipulations with quantitative and qualitative evaluations of its robustness, capacity and accuracy. In addition, our method generalizes 3D watermarking use cases. We further analyze the efficiency of different architectures and explore the performance of our framework on several real-world renderers. 

It is worth noting that more work needs to be done to make it more applicable to real-world cases.
For example, our decoder needs to be re-trained for totally different types of rendering techniques. Improving and generalizing our decoder is an interesting research topic. For future work, the robustness to non-differentiable 3D distortions and making the decoder to be able to retrieve messages encoded by traditional encoders are worth studying. Another direction is to explore better differentiable renderers, so that the framework can simulate and generalize to more complex 3D models or shading effects.

\section{Supplementary Materials}
\label{sec:supp}
The detailed architectures of our encoders and decoder are in Sec.~\ref{ssec:supp_archs}. The parameters and implementation details are in Sec.~\ref{ssec:supp_diffren}.
Thorough evaluations of each distortion are in Sec.~\ref{ssec:supp_dist}.
Lastly, more results generated by our watermark encoders and the differences are in Sec.~\ref{ssec:supp_results}.

\subsection{Architectures}
\label{ssec:supp_archs}

As described in the paper, we used variations of PointNet~\cite{PointNet} as backbone architectures for the vertex encoder network.
For the texture encoder, we use CNN-based architectures such as HiDDeN~\cite{HiDDeN}'s encoder, or a fully convolutional U-Net~\cite{UNet}.

\subsubsection{3D Vertex Encoder}

\begin{figure}[h]
  \centering
  \includegraphics[width=1.0\linewidth]{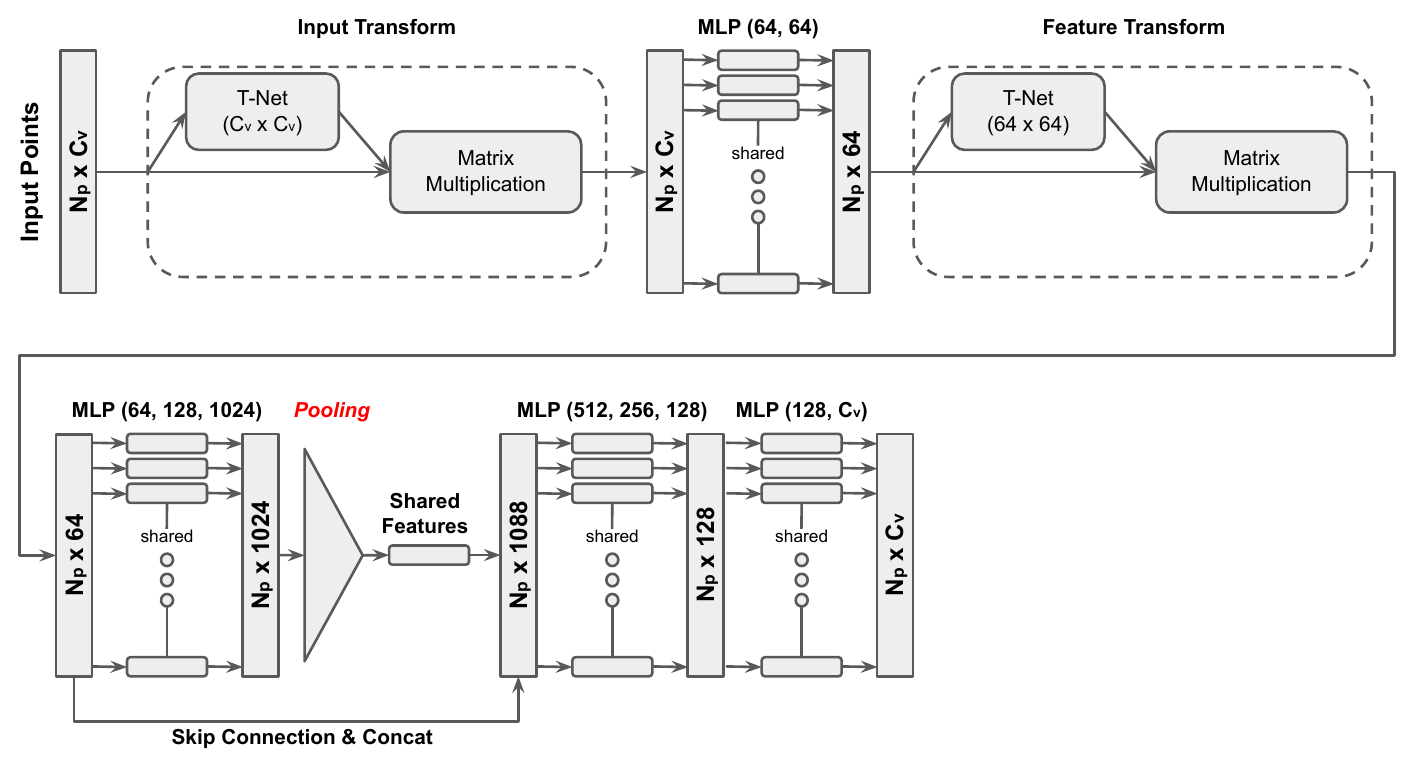}
  \caption{PointNet architecture}
  \label{fig:supp_pointnet_arch}
\end{figure}

\textbf{PointNet}
Fig.~\ref{fig:supp_pointnet_arch} shows the PointNet architecture,~\emph{i.e.}, our encoder backbone network.
Most parts are similar to~\cite{PointNet}, yet, there are a few differences: 1) we changed the shape of input points which originally only accept 3D positional element, $\{x, y, z\}$, and now could accept any numbers of vertex elements $C_v$; 2) the pooling layer (marked as red in Fig.~\ref{fig:supp_pointnet_arch}) in the architecture can be either max pooling (PointNet) or global pooling (PointNet v2 in the paper) for accepting mesh inputs with different size.

\subsubsection{2D Texture Encoder}
\label{ssec:supp_tex_encoder}

\textbf{CNN Encoder}
Our CNN encoder is based on HiDDeN~\cite{HiDDeN}'s encoder architecture.
The input texture is first passed by four CBR blocks containing $3 \times 3$ convolution, batch normalization, and ReLU activation. Each block has 64 units per layer.
Then, input messages are repeated to have a same dimensions of $height \times width$ of the input textures, and concatenated in the channel dimension.
The message appended feature maps are further connected to two CBR blocks.
The last CBR block has the unit of $3$ so that the output has the same shape of input texture (which is the message embedded texture).
For the convolutional layers in the encoder architecture, we use `VALID` padding.

\textbf{U-Net}
U-Net~\cite{UNet} architecture systematically combines the autoencoder architecture and skip-connection scheme as shown in~\cite{ResNet}.
For our texture encoder, we modified U-Net architecture to make it fully convolutional through: 1) removing the fully-connected layer which generates the latent space vector in between the autoencoder's encoder and decoder; 2) using $\{64, 128, 256, 512\}$ units per U-Net block with max pooling by $2$ in each block.

\subsubsection{Image Decoder}

\textbf{CNN Decoder}
Our CNN decoder is based on HiDDeN~\cite{HiDDeN}'s decoder architecture.
The base CBR block is same as the CNN encoder as described in Sec.~\ref{ssec:supp_tex_encoder}.
We use seven CBR blocks. The last two CBR blocks are applying stride with $2$.
Then, global pooling is applied to the last CBR block to accept any image dimensions.
Lastly, a fully-connected layer is used to generate the output message logits. To have the same dimension of message bits, the output unit of fully-connected layer is $N_b$.

\subsection{Differentiable Rendering}
\label{ssec:supp_diffren}
As mentioned in the paper, we leverage the state-of-the-art work in differentiable rendering and follow the work by Genova et al.~\cite{genova2018unsupervised}. We explain the steps in more details here:

\begin{itemize}
    \item The differentiable renderer first takes $N_v \times 3$ world-space vertices and a sampled camera position as input, and computes $N_v \times 4$ projected vertices using camera projection $\mathcal{P}$, following OpenGL convention~\cite{kessenich2016opengl}.
    \item Then it rasterizes the triangles by identifying the front-most triangle ID $\mathcal{T}$ at each pixel and computing the barycentric coordinates of the pixel inside triangles $\mathcal{B}$. 
    \item After rasterization, we interpolate $N_v \times 5$ vertex attributes, i.e. normals and uvs, at the pixels using the barycentric coordinates $\mathcal{B}$ and triangle IDs $\mathcal{T}$. We take the interpolated per-vertex attributes and the lighting parameters $\mathcal{L}$ to compute the shaded colors $\mathcal{C}$. Here we use a Phong model with parameters $k_a=0.8$, $k_d=1.4$ and $k_r=0$, and set the constant term $K_c=1.0$, the linear term $K_l=0.07$, and the quadratic term $K_q=0.017$ to calculate attenuation value.
    \item Finally, we form a $h \times w \times 4 $ buffer of per-pixel clip-space positions $\mathcal{V}$, then apply perspective division and viewport transformation to produce a $h \times w \times 2$ screen-space splat position buffer $\mathcal{S}$. In our implementation, the rendered height $h$ and $w$ is 400 and 600.
\end{itemize}

\subsection{Evaluations of Distortions}
\label{ssec:supp_dist}

\begin{figure*}[ht]
  \centering
  \begin{subfigure}{.08\linewidth}
    \centering
    Noise
  \end{subfigure}
  \begin{subfigure}{0.45\linewidth}
    \centering
    \includegraphics[width=1.0\linewidth]{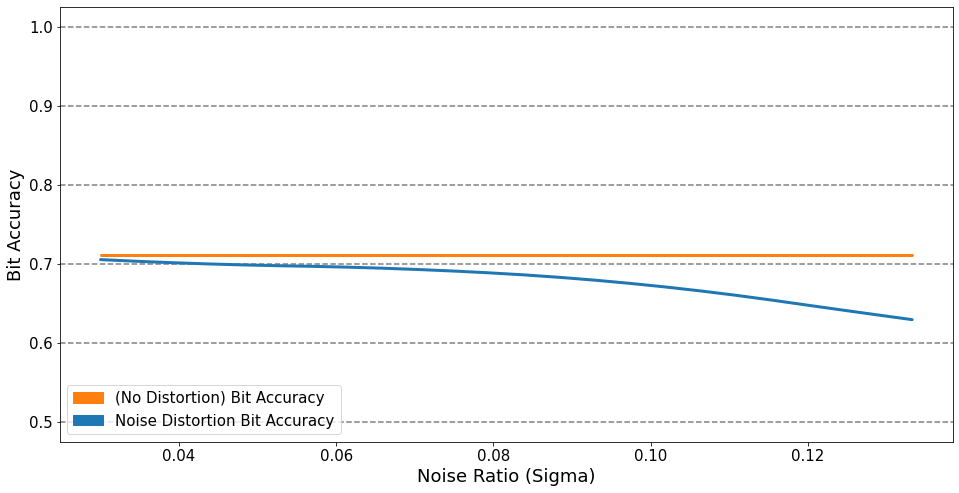}
  \end{subfigure}
  \begin{subfigure}{0.45\linewidth}
    \centering
    \includegraphics[width=1.0\linewidth]{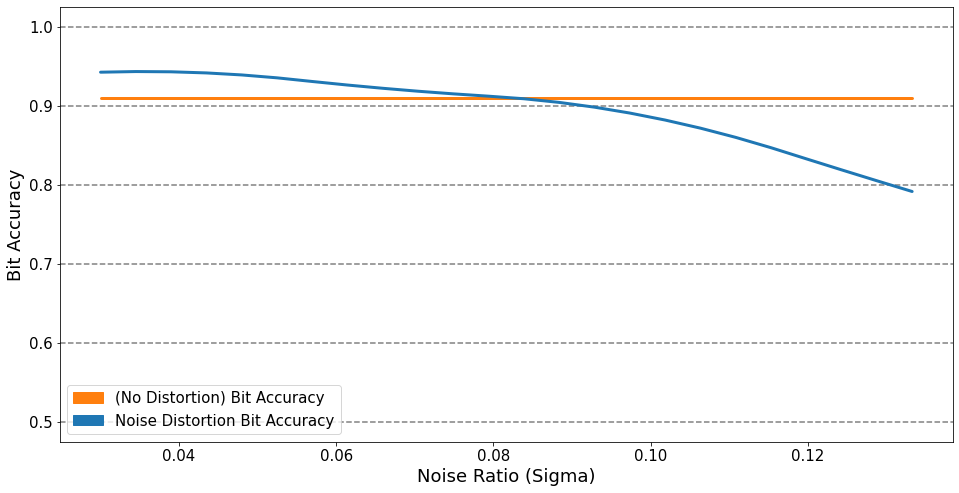}
  \end{subfigure}
  \\
  \begin{subfigure}{.08\linewidth}
    \centering
    Rotation
  \end{subfigure}
  \begin{subfigure}{0.45\linewidth}
    \centering
    \includegraphics[width=1.0\linewidth]{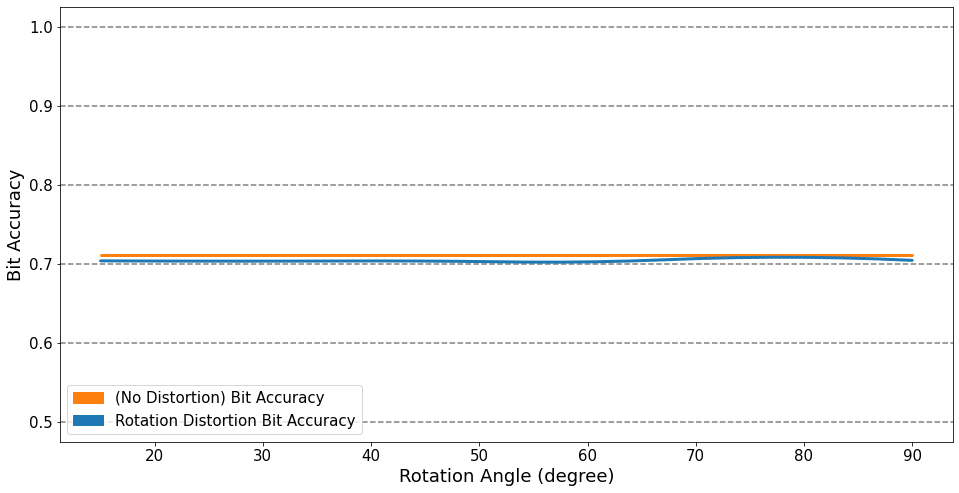}
  \end{subfigure}
  \begin{subfigure}{0.45\linewidth}
    \centering
    \includegraphics[width=1.0\linewidth]{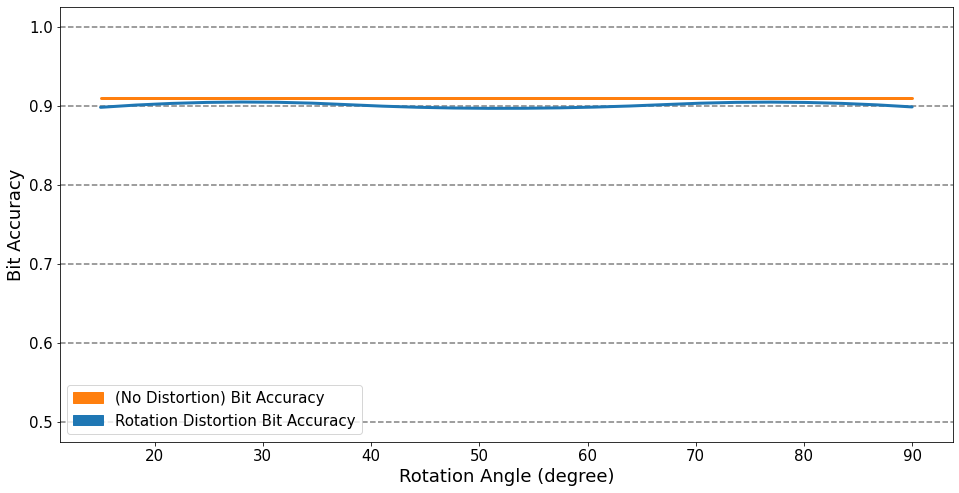}
  \end{subfigure}
  \\
  \begin{subfigure}{.08\linewidth}
    \centering
    Scaling
  \end{subfigure}
  \begin{subfigure}{0.45\linewidth}
    \centering
    \includegraphics[width=1.0\linewidth]{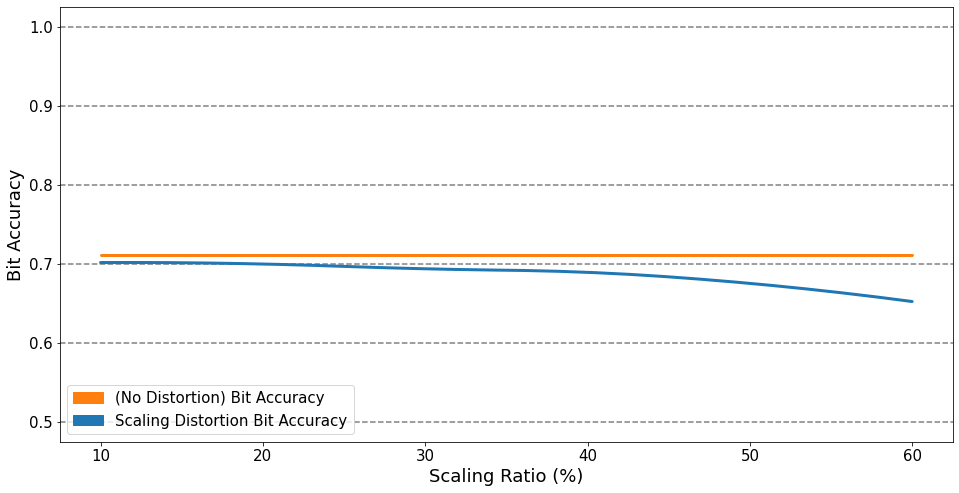}
  \end{subfigure}
  \begin{subfigure}{0.45\linewidth}
    \centering
    \includegraphics[width=1.0\linewidth]{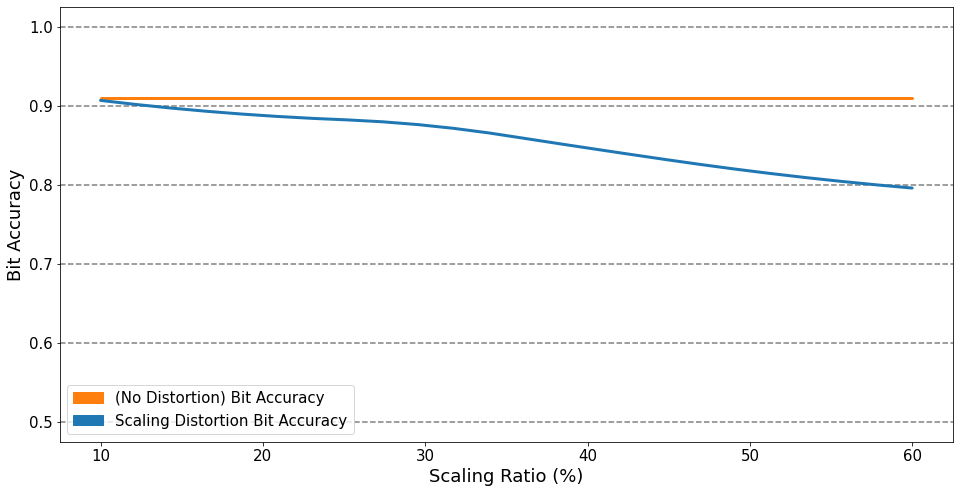}
  \end{subfigure}
  \\
  \begin{subfigure}{.08\linewidth}
    \centering
    Cropping
  \end{subfigure}
  \begin{subfigure}{0.45\linewidth}
    \centering
    \includegraphics[width=1.0\linewidth]{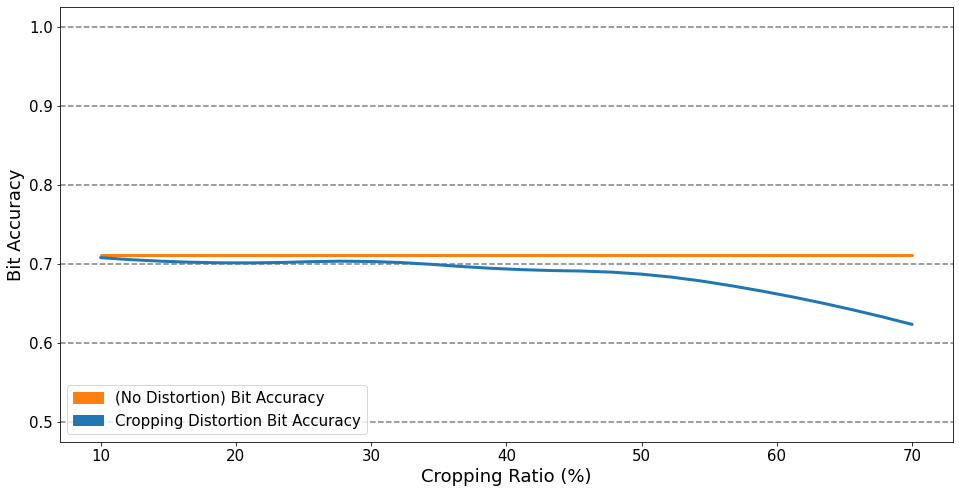}
    \caption{Real Texture}
  \end{subfigure}
  \begin{subfigure}{0.45\linewidth}
    \centering
    \includegraphics[width=1.0\linewidth]{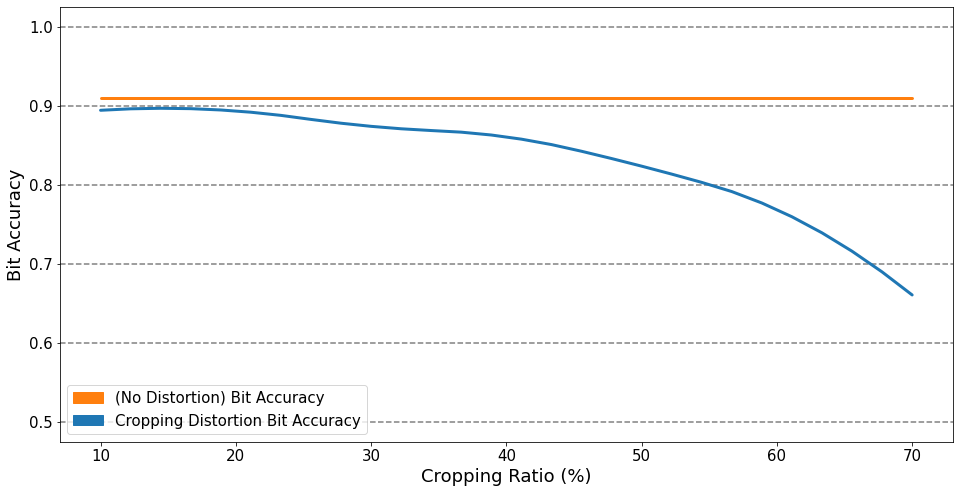}
    \caption{Noise Texture}
  \end{subfigure}
  \caption{Bit accuracy against distortion strength.}
  \label{fig:supp_distortions}
\end{figure*}

We verified the robustness of our networks for distortions. Our network is trained with four distortions: additive noise, scaling, rotation, and cropping.
As described in the paper, we trained our network with two different types of textures: real and noise textures.
Fig.~\ref{fig:supp_distortions} shows the graphs between bit accuracy and distortion strength.
The results are trained with message length $8$, real textures (left) and noise textures (right).
Note that noise distortion plots in Fig.~\ref{fig:supp_distortions} are based on $\mu=\{\pm0.1, \pm0.15, \pm0.2, \pm0.25, \pm0.3, \pm0.4\}$, and $\sigma=\{0.03, 0.05, 0.06, 0.0833, 0.1, 0.133 \}$.
As we can see in the Fig.~\ref{fig:supp_distortions}, the bit accuracy for network trained with real textures is lower than noise textures, however, the tendencies between bit accuracy and distortion strength are similar.
Overall, our network is robust to noise, rotation, scaling, and cropping distortions.

\begin{figure}[h]
  \centering
  \begin{subfigure}{0.32\linewidth}
    \centering
    \includegraphics[width=1.0\linewidth]{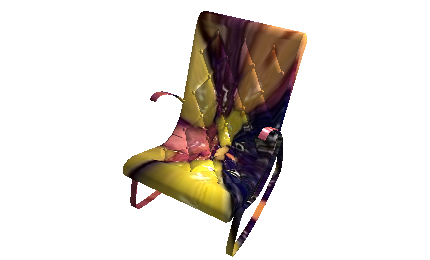}
  \end{subfigure}
  \begin{subfigure}{0.32\linewidth}
    \centering
    \includegraphics[width=1.0\linewidth]{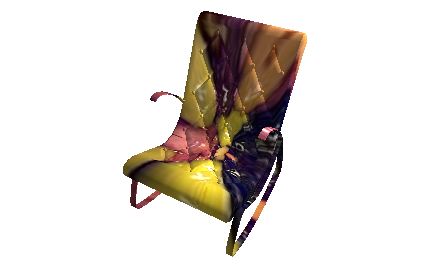}
  \end{subfigure}
  \begin{subfigure}{0.32\linewidth}
    \centering
    \includegraphics[width=1.0\linewidth]{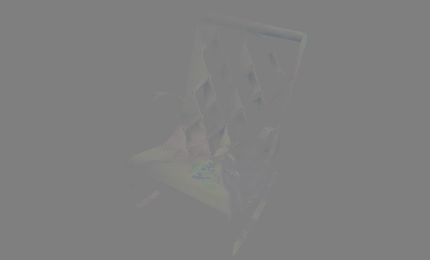}
  \end{subfigure}
  \\
  \begin{subfigure}{0.32\linewidth}
    \centering
    \includegraphics[width=1.0\linewidth]{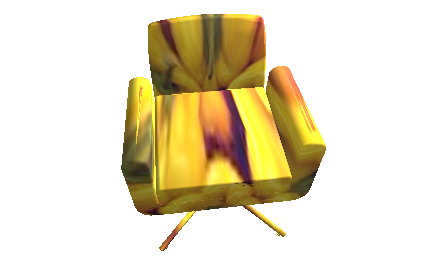}
  \end{subfigure}
  \begin{subfigure}{0.32\linewidth}
    \centering
    \includegraphics[width=1.0\linewidth]{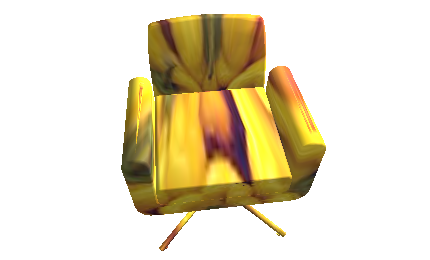}
  \end{subfigure}
  \begin{subfigure}{0.32\linewidth}
    \centering
    \includegraphics[width=1.0\linewidth]{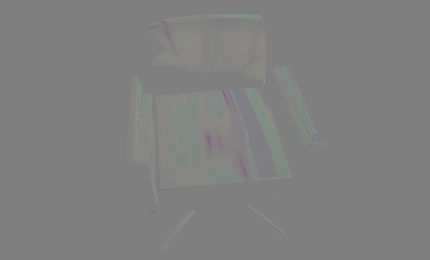}
  \end{subfigure}
  \\
  \begin{subfigure}{0.32\linewidth}
    \centering
    \includegraphics[width=1.0\linewidth]{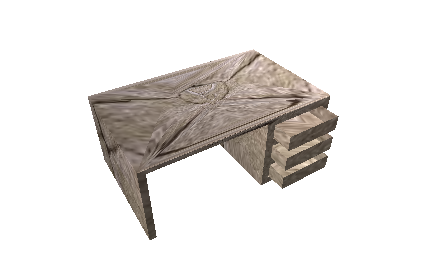}
  \end{subfigure}
  \begin{subfigure}{0.32\linewidth}
    \centering
    \includegraphics[width=1.0\linewidth]{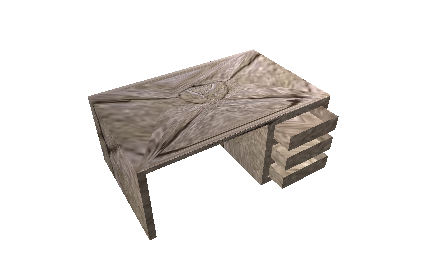}
  \end{subfigure}
  \begin{subfigure}{0.32\linewidth}
    \centering
    \includegraphics[width=1.0\linewidth]{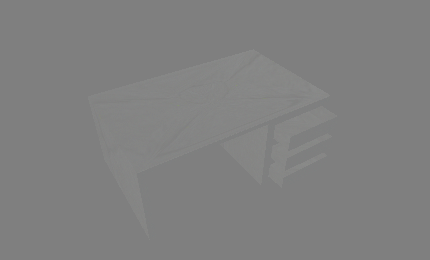}
  \end{subfigure}
  \\
  \begin{subfigure}{0.32\linewidth}
    \centering
    \includegraphics[width=1.0\linewidth]{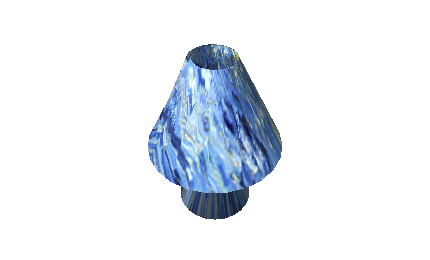}
    \caption{Not Watermarked}
  \end{subfigure}
  \begin{subfigure}{0.32\linewidth}
    \centering
    \includegraphics[width=1.0\linewidth]{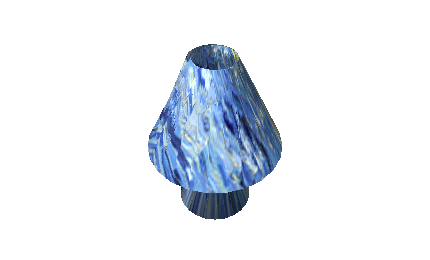}
    \caption{Watermarked}
  \end{subfigure}
  \begin{subfigure}{0.32\linewidth}
    \centering
    \includegraphics[width=1.0\linewidth]{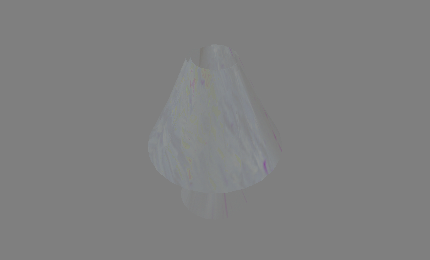}
    \caption{Difference (4x)}
  \end{subfigure}
  \caption{The rendered images from input meshes (a), watermarked meshes (b) , and the difference images (c), with real textures.}
  \label{fig:supp_rendered_images_real}
\end{figure}

\begin{figure}[h]
  \centering
  \begin{subfigure}{0.32\linewidth}
    \centering
    \includegraphics[width=1.0\linewidth]{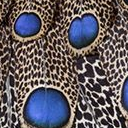}
  \end{subfigure}
  \begin{subfigure}{0.32\linewidth}
    \centering
    \includegraphics[width=1.0\linewidth]{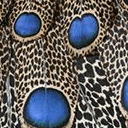}
  \end{subfigure}
  \begin{subfigure}{0.32\linewidth}
    \centering
    \includegraphics[width=1.0\linewidth]{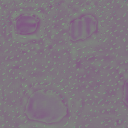}
  \end{subfigure}
  \\
  \begin{subfigure}{0.32\linewidth}
    \centering
    \includegraphics[width=1.0\linewidth]{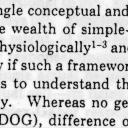}
  \end{subfigure}
  \begin{subfigure}{0.32\linewidth}
    \centering
    \includegraphics[width=1.0\linewidth]{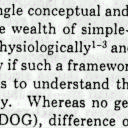}
  \end{subfigure}
  \begin{subfigure}{0.32\linewidth}
    \centering
    \includegraphics[width=1.0\linewidth]{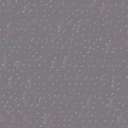}
  \end{subfigure}
  \\
  \begin{subfigure}{0.32\linewidth}
    \centering
    \includegraphics[width=1.0\linewidth]{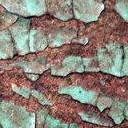}
    \caption{Not Watermarked}
  \end{subfigure}
  \begin{subfigure}{0.32\linewidth}
    \centering
    \includegraphics[width=1.0\linewidth]{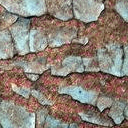}
    \caption{Watermarked}
  \end{subfigure}
  \begin{subfigure}{0.32\linewidth}
    \centering
    \includegraphics[width=1.0\linewidth]{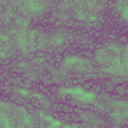}
    \caption{Difference (4x)}
  \end{subfigure}
  \caption{Comparisons between not watermarked and watermarked real textures.}
  \label{fig:supp_texture_real}
\end{figure}

\subsection{More Results}
\label{ssec:supp_results}
We provide more rendered results that rendered from original meshes, watermarked meshes, and the differences between the two images (Fig.~\ref{fig:supp_rendered_images_real}). Also original textures, watermarked textures, and the difference are shown in Fig.~\ref{fig:supp_texture_real}.

{\small
\bibliographystyle{ieee_fullname}
\bibliography{main}
}

\end{document}